%% file: root.tex
\documentclass[letterpaper, 10 pt, conference]{ieeeconf}  %
\usepackage{tipa}
\usepackage{bm}
\usepackage{url}
\usepackage{balance}
\usepackage{graphicx}
\usepackage{amsfonts}
\usepackage{fancyhdr}
\usepackage{comment}
\usepackage{times}
\usepackage{amsmath}
\usepackage{changepage}
\usepackage{amssymb}
\usepackage{enumerate}
\usepackage{algorithmic}
\usepackage{bm}
\usepackage{subfigure}
\usepackage{calligra}
\usepackage{multirow}
\usepackage{tabularx}
\usepackage{booktabs}
\usepackage{mathtools}
\usepackage{arydshln}
\usepackage{latexsym}
\usepackage{amsmath}
\usepackage{amssymb}
\usepackage{color}
\usepackage{soul}
\usepackage{gensymb}
\usepackage{cite}
\usepackage{color}
\usepackage{cuted}
\usepackage{etoolbox}
\usepackage{capt-of}
\usepackage{subfigure}
\usepackage{ulem}
\usepackage{float}
\usepackage[ruled,vlined]{algorithm2e}
\makeatletter
\def\hlinewd#1{%
  \noalign{\ifnum0=`}\fi\hrule \@height #1 \futurelet
   \reserved@a\@xhline}
\makeatother
\AfterEndEnvironment{strip}{\leavevmode}

\newcommand{\redtext}{\textcolor[rgb]{1,0,0}}
\IEEEoverridecommandlockouts
\title{\LARGE \bf
MLOD: Awareness of Extrinsic Perturbation in Multi-LiDAR 3D Object Detection for Autonomous Driving
\author{Jianhao Jiao$^{*}$, Peng Yun$^{*}$, Lei Tai, Ming Liu%
}
\thanks{This work was supported by the National Natural Science Foundation of China, under grant No. U1713211, Collaborative Research Fund by Research Grants Council Hong Kong, under Project C4063-18G, and the Research Grant Council of Hong Kong SAR Government, China, under Project No. 11210017, awarded to Prof. Ming Liu.}
\thanks{$*$ Equal contribution (alphabet order). Jianhao Jiao, Peng Yun, Ming Liu are with the Robotics Institute, Intelligent Autonomous Driving Center, The Hong Kong University of Science and Technology, Hong Kong SAR, China.
{\tt\small \{jjiao, pyun, eelium\}}@ust.hk. 
Lei Tai is with IAS BU, Huawei Technologies. {\tt\small \{tailei2\}}@huawei.com.}
}
\begin{document}
\maketitle

\begin{abstract}
    Extrinsic perturbation always exists in multiple sensors.
    In this paper, we focus on the extrinsic uncertainty in multi-LiDAR systems for 3D object detection.
    We first analyze the influence of extrinsic perturbation on geometric tasks with two basic examples.
	To minimize the detrimental effect of extrinsic perturbation,
	we propagate an uncertainty prior on each point of input point clouds, and use this information to boost an approach for 3D geometric tasks.
    Then we extend our findings to propose a multi-LiDAR 3D object detector called \textit{MLOD}. \textit{MLOD} is a two-stage network where the multi-LiDAR information is fused through various schemes in stage one, and the extrinsic perturbation is handled in stage two.
	We conduct extensive experiments on a real-world dataset, and demonstrate both the accuracy and robustness improvement of \textit{MLOD}.
	The code, data and supplementary materials are available at: \url{https://ram-lab.com/file/site/mlod}.
\end{abstract}

\input{introduction}

\input{related_work}

\input{problem}

\input{uncertainty_intro}

\input{methodology}

\input{experiment}

\input{conclusion}

\balance
\bibliographystyle{IEEEtran}
\bibliography{reference}

\end{document}

%% file: introduction.tex
\section{Introduction}
\label{sec:introduction}

3D object detection is a fundamental module in robotic systems.
As the front-end of a system, it enables vehicles to recognize key objects such as cars and pedestrians, which is indispensable for high-level decision making.
Compared with camera-based detectors,
the LiDAR-based detectors perform better in several challenging scenarios because of their activeness as sensors and distance measurement ability for surroundings. 
However, LiDARs commonly suffer from data sparsity and a limited vertical field of view 
\cite{jiao2019automatic}.
For instance, LiDARs' points distribute loosely, which induces a mass of empty regions between two nearby scans.
In this paper, we consider a multi-LiDAR system, a setup with excellent potential to solve 3D object detection.
Compared with single-LiDAR setups, multi-LiDAR systems enable a vehicle to maximize its perceptual awareness of environments and obtain sufficient measurements.
The decreasing price of LiDARs also makes such systems accessible to many modern self-driving cars \cite{jiao2019automatic,sun2020scalability,lyft2019}.

\begin{figure}[]
	\centering
	\subfigure[Multiple point clouds are merged with little extrinsic perturbation.]
	{\centering\includegraphics[width=0.45\textwidth]{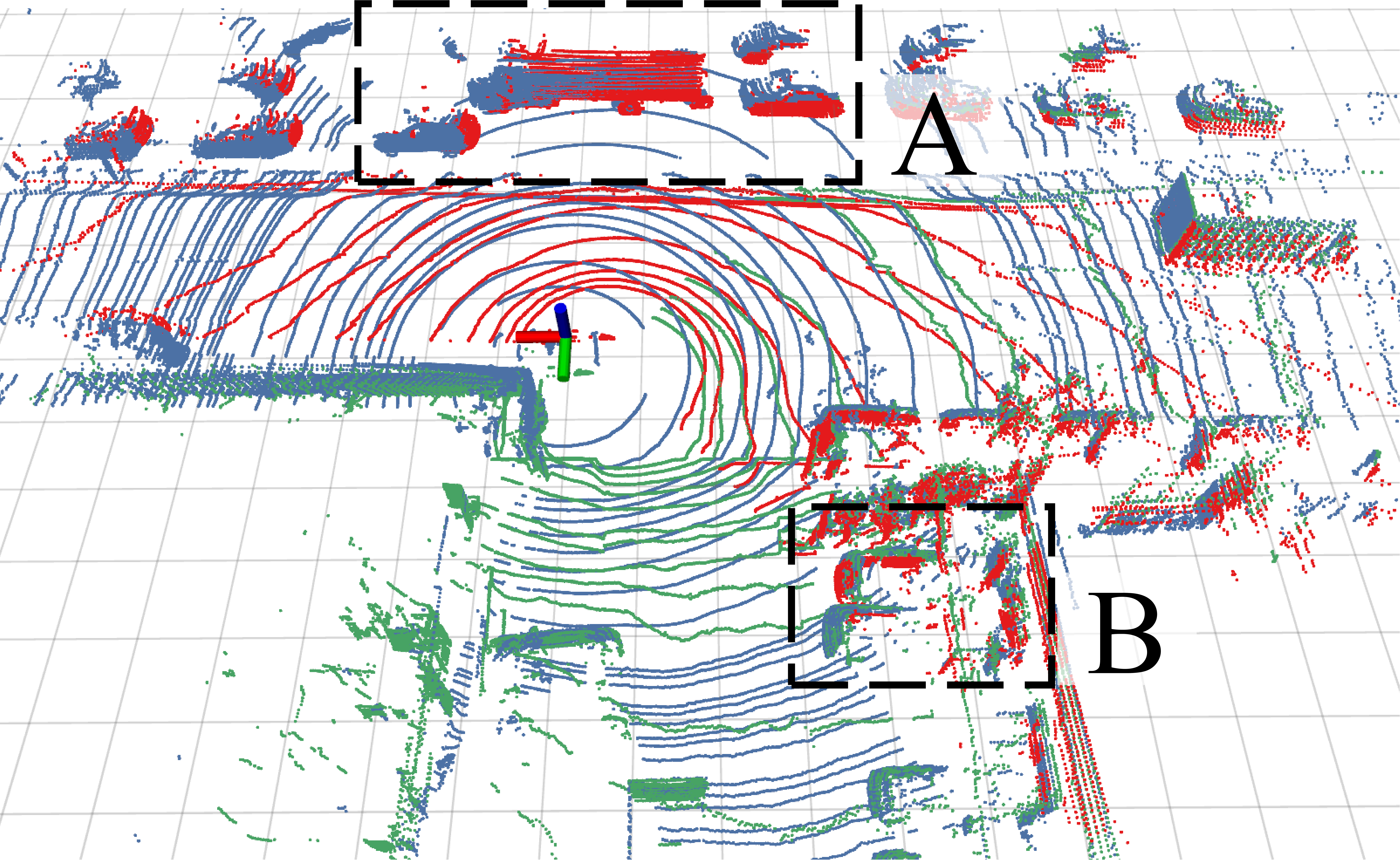}}
	\hspace{0.25cm}
	\subfigure[The details in the region A and B of the point cloud.]
	{\centering\includegraphics[width=0.45\textwidth]{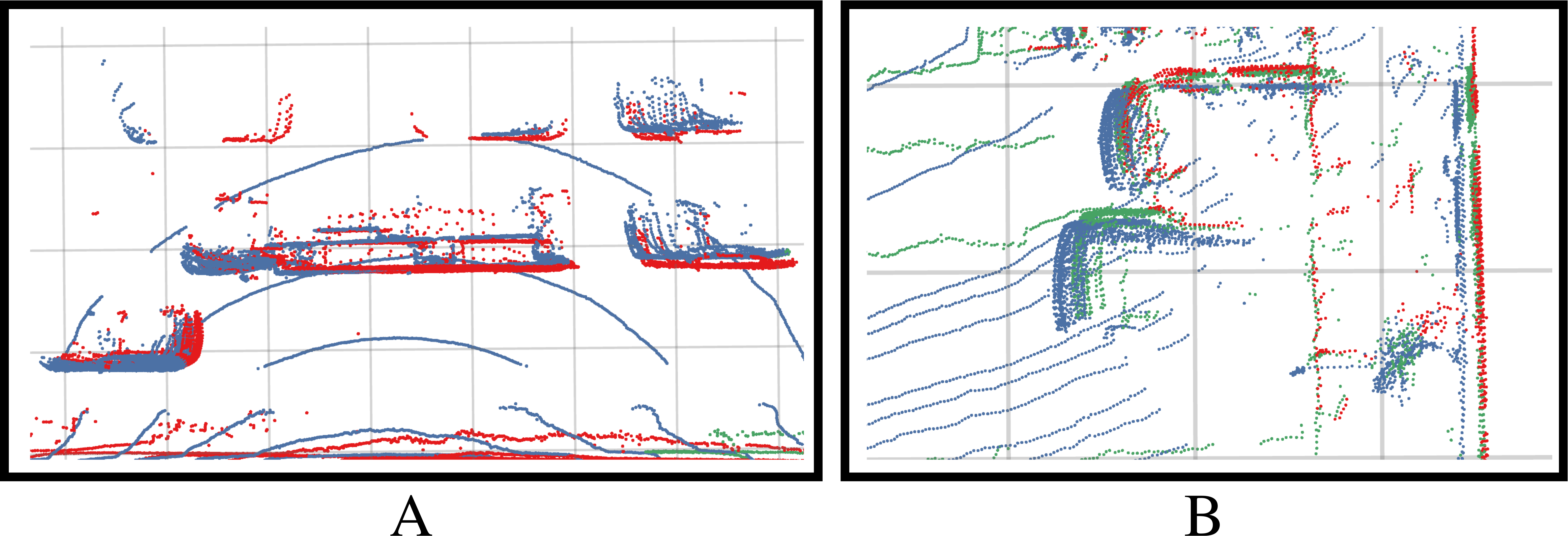}} 
	\caption{
		(a) A point cloud example from the LYFT dataset \cite{lyft2019}. It is merged by transforming point clouds perceived by top (blue), front-left (green), and front-right (red) LiDARs into the base frame. 
		We simulate perturbation on the rotation of $1^{\circ}$ in yaw.
		(b) The details in the region A and B with a zooming effect. Massive noisy points are induced by extrinsic perturbation, while they should be sharp or flat in perturbation-free situations.}
	\label{fig.introduction_perturbation}    
	\vspace{-0.7cm}
\end{figure}

However, two challenges affect the development of multi-LiDAR object detection. 
One difficulty is multi-LiDAR fusion. As described in \cite{chen2017multi}, current data fusion methods all have pros and cons.
It requires us to perform extensive efforts on possible fusion schemes in real tests for an excellent detection algorithm.
Another issue is that extrinsic perturbation is inevitable for sensor fusion during long-term operation because of factors such as vibration, temperature drift, and calibration error \cite{ling2016high}. 
Especially, wide baseline stereo cameras or vehicle-mounted multi-LiDAR systems suffer even more extrinsic deviations than the normal one.
Even though online calibration has been proposed to handle this issue \cite{lin2020decentralized}, life-long calibration is always challenging \cite{maddern20171}. 
These methods typically require environmental or motion constraints with full observability. 
Otherwise, the resulting extrinsics may become suboptimal or unreliable.
As identified in both Fig. \ref{fig.introduction_perturbation} and Section \ref{sec.problem_propagation}, extrinsic perturbation is detrimental to the measurement accuracy even with a small change.
But this fact is often neglected by the research community.

To tackle the challenges, we propose a \textbf{M}ulti-\textbf{L}iDAR \textbf{O}bject \textbf{D}etector (\textit{MLOD}) to predict states of objects on point clouds.
We design a two-stage procedure to estimate 3D bounding boxes, and demonstrate two innovations in \textit{MLOD}.
First, we explore three fusion schemes to exploit multi-view point clouds in a stage-1 network for generating object proposals. These schemes perform LiDAR fusion at the input, feature, and output phases, respectively.
Second, we develop a stage-2 network to handle extrinsic perturbation and refine the proposals.
The experimental results demonstrate that \textit{MLOD} outperforms single-LiDAR detectors up to 9.7AP in perturbation-free cases.
When considering extrinsic perturbation, \textit{MLOD} consistently improves the performance of its stage-1 counterpart.
To the best of our knowledge, this is the first work to systematically study the multi-LiDAR object detection with consideration of extrinsic perturbation.
Overall, our work has the following contributions:
\begin{enumerate}
	\item We analyze the influence of extrinsic perturbation on multi-LiDAR-based geometric tasks, and also demonstrate that the usage of input uncertainty prior improves the robustness of an approach against such effect. 
	\item We propose a unified and effective two-stage approach for multi-LiDAR 3D object detection, where the multi-LiDAR information is fused in the first stage, and extrinsic perturbation is handled in the second stage.
	\item We exhaustively evaluate the proposed approach in terms of accuracy and robustness under extrinsic perturbation on a real-world self-driving dataset.
\end{enumerate}

%% file: related_work.tex
\section{Related Work}
\label{sec:related_work}

We briefly review methods on LiDAR-based object detection and uncertainty estimation of deep neural networks.

\subsection{3D Object Detection on Point Clouds}
The LiDAR-based object detectors are generally categorized into grid-based \cite{li20173d,zhou2018voxelnet,yan2018second} and point-based methods \cite{qi2017pointnet,qi2017pointnet++,shi2019pointrcnn}.
3D-FCN \cite{li20173d} implemented 3D volumetric CNN on voxelized point clouds. 
But it commonly suffers a high computation cost due to dense convolution on sparse point clouds. VoxelNet \cite{zhou2018voxelnet} was proposed as an end-to-end network to learn features.
However, operation on grids is inefficient since LiDAR's points are sparse. 
To cope with this drawback, SECOND \cite{yan2018second} utilized the spatially sparse convolution \cite{graham2017submanifold} 
to replace the 3D dense convolution layers.
In this paper, we adopt SECOND as our basic proposal generator and propose three schemes for multi LiDAR fusion in the first stage.
Qi et al. proposed PointNet \cite{qi2017pointnet} to directly learn features from raw data with a symmetric function. And their follow-up work presented a set-abstraction block to capture local features \cite{qi2017pointnet++}.
PointRCNN \cite{shi2019pointrcnn} exploits PointNet to learn point-wise features and segments foreground for 
autonomous driving. 
In this paper, we extend PointNet to design the stage-2 network of \textit{MLOD} with an awareness of extrinsic perturbation.

Researchers also explored fusion methods that further exploit images to improve the LiDAR-based approaches.
For example, MV3D \cite{chen2017multi} combined features from multiple views, including a bird-eye view and front view, to conduct classification and regression.
Several methods \cite{qi2018frustum, xu2018pointfusion} separate the process of detection into two stages. They first projected region proposals generated by the camera-based detectors into 3D space, and then used PointNets to segment and to classify objects on point clouds.
But these methods are aimed at data fusion from cameras and LiDARs and assume sensors to be well-calibrated.
In contrast, our work focuses on multi-LiDAR fusion, and tries to minimize the negative effect of extrinsic perturbation for 3D object detection.

\subsection{Uncertainty Estimation in Object Detection}
The problem of uncertainty estimation is essential to the reliability of an algorithm, and has attracted much attention in recent years. 
As the two main types of uncertainties in deep neural networks: aleatoric and epistemic uncertainty, they are explained in \cite{kendall2017uncertainties, kendall2017bayesian}.
In \cite{kendall2017bayesian}, the authors also demonstrated the benefits of modeling uncertainties for vision tasks.
As an extension, Feng et al. \cite{feng2019leveraging} proposed an approach to capture uncertainties for 3D object detection.
Generally, the data noise is modeled as a unique Gaussian variable in these works.
However, the sources of data uncertainties in multi-LiDAR systems are much more complicated.
In this paper, we take both the measurement noise and extrinsic perturbation into account, and analyze their effect on multi-LiDAR-based object detection.
Furthermore, we propagate the Gaussian uncertainty prior of both the extrinsics and measurement to model the input data uncertainty. 
This additional cue is utilized to improve the robustness of \textit{MLOD} against extrinsic perturbation.

%% file: problem.tex
\begin{table}[]
	\centering
	\caption{Nomenclature}
	\renewcommand\arraystretch{1.1}
	\begin{tabularx}{0.47\textwidth}{cX}
		\hline
		\toprule[0.03cm]
		Notation & Explanation \\ 
		\hline
		\toprule[0.01cm]
		$\{\}^{b}, \{\}^{l^{i}}$ & Frame of the base and the $i^{th}$ LiDAR.\\
		$\mathcal{P}^{l^{i}}$ & Raw point cloud captured by the $i^{th}$ LiDAR.\\	
		$\mathcal{B}$ & Set of estimated 3D bounding boxes.\\		
		$\mathbf{T}$ & Transformation matrix in the \textit{Lie group} $SE(3)$.\\		
		$\mathbf{R}$ & Rotation matrix in the \textit{Lie group} $SO(3)$.\\				
		$\mathbf{t}$ & Translation vector in $\mathbb{R}^{3}$.\\						
		$\bm{\Theta}$ & Measurement and extrinsic uncertainty prior.\\
		$\bm{\Xi}$ & Associated covariance of each point.\\	
		$\alpha$ & Scaling parameter of $\bm{\Theta}$.\\
		\hline
		\toprule[0.03cm]		
	\end{tabularx}
	\label{tab.nomenclature}
	\vspace{-0.3cm}
\end{table}

\section{Notations and Problem Statement}
\label{sec.notations_and_problem_statement}

The nomenclature is shown in Tab. \ref{tab.nomenclature}.
The transformation from $\{\}^{b}$ to $\{\}^{l^{i}}$ is denoted by $\mathbf{T}^{b}_{{l}^{i}}$. 
Our perception system consists of one primary LiDAR which is denoted by $l^{1}$ and multiple auxiliary LiDARs.
The primary LiDAR defines the base frame and the auxiliary LiDAR provides an additional field of view (FOV) and measurements to alleviate the occlusion problem and the sparsity drawback of the primary LiDAR.
Extrinsics describe the relative transformation from the base frame to frames of auxiliary LiDARs. 
With the extrinsics, all measurements or features from different LiDARs are transformed into the base frame for data fusion. 
LiDARs are assumed to be synchronized that multiple point clouds are perceived at the same time.

In this paper, we focus on 3D object detection with a multi-LiDAR system. 
The extrinsic perturbation is also considered, which indicates the small but unexpected change on transformation from the base frame to other frames over time.
Our goal is to estimate a series of 3D bounding boxes covering objects of predefined classes.
Each bounding box $\mathbf{b}\in\mathcal{B}$ is parameterized as $[c, x,y,z,w,l,h,\gamma]$, $c$ is the class of a bounding box, $[x,y,z]$ denotes a box's bottom center, 
$[w,h,l]$ represent the sizes along the $x\textendash$, $y\textendash$, 
and $z\textendash$ axes respectively,
as well as $\gamma$ for the rotation of the 3D bounding box along the $z\textendash$ axis in the range of $(0, \pi]$.

%% file: uncertainty_intro.tex
\section{Propagating Extrinsic Uncertainty on Points}
\label{sec.problem}

We begin with providing preliminaries about the uncertainty representation and propagation. We then use an example to demonstrate the negative effect on points caused by extrinsic perturbation. Finally, we also conduct a plane fitting experiment to show that the extrinsic covariance prior can be utilized to make a fitting approach robust.

\subsection{Preliminaries}
We employ the method in \cite{barfoot2014associating} to represent the uncertainty. 
For convenience, rotation and translation are used to indicate a transformation.
We first define a random variable for $\mathbb{R}^{3}$ with small perturbation according to
\begin{equation}
	\label{equ.translation_with_uncertainty}
    \mathbf{t}
    =
    \bm{\rho}+\bar{\mathbf{t}},
    \ \ \ 
    \bm{\rho}\sim\mathcal{N}(\mathbf{0}, \mathbf{P}),
\end{equation}
where $\bar{\mathbf{t}}$ is a noise-free translation and $\bm{\rho}\in\mathbb{R}^{3}$ is a zero-mean Gaussian variable with covariance $\mathbf{P}$.
We can also define a rotation for $SO(3)$ as\footnote{The $\wedge$ operator turns a $3\times1$ vector into the corresponding skew symmetric matrix in the \textit{Lie algebra} $\mathfrak{so}(3)$. The exponential map $\exp$ associates an element of $\mathfrak{so}(3)$ to a rotation in $SO(3)$. The closed-form expression for the matrix exponential can be found in \cite{barfoot2014associating}.}
\begin{equation}
\label{equ.rotation_with_uncertainty}
\mathbf{R} = \exp(\bm{\phi}^{\wedge})\bar{\mathbf{R}}, 
\ \ \ \bm{\phi}\sim\mathcal{N}(\mathbf{0},\bm{\Phi}),
\end{equation}
where $\bar{\mathbf{R}}$ is a noise-free rotation and $\bm{\phi}\in\mathbb{R}^{3}$ is a small zero-mean Gaussian variable with the covariance $\bm{\Phi}$. 
With \eqref{equ.rotation_with_uncertainty}, we can represent a noisy transformation by storing the mean as $[\bar{\mathbf{t}}, \bar{\mathbf{R}}]$ and using $[\bm{\rho}, \bm{\phi}]$ for perturbation on the vector space. 
Similarly,
a point with the perturbation in $\mathbb{R}^{3}$ is written as
\begin{equation}
\mathbf{p}
=
\bm{\zeta}
+
\bar{\mathbf{p}}
,
\ \ \   
\bm{\zeta}\sim\mathcal{N}(\mathbf{0},\textbf{Z}),
\end{equation}
where $\bm{\zeta}$ is zero-mean Gaussian with the covariance $\textbf{Z}$.
With the above representations, we can pass the Gaussian representation of a point through a noisy rotation and translation to produce a mean and covariance for its new measurement. 
Here, we can use $[\bar{\mathbf{t}}^{b}_{l^{i}}, \bar{\mathbf{R}}^{b}_{l^{i}}]$ to indicate the ground-truth extrinsics of a multi-LiDAR system, and $[\bm{\rho}, \bm{\phi}, \bm{\zeta}]$ to indicate both the extrinsic and measurement perturbation.
By transforming $\mathbf{p}\in\mathcal{P}^{i}$ into $\{\}^{b}$, we have
\begin{equation}
\begin{aligned}
\mathbf{y}
\triangleq
\mathbf{R}^{b}_{l^{i}}\mathbf{p} + \mathbf{t}^{b}_{l^{i}}
&=
\exp(\bm{\phi}^{\wedge})\bar{\mathbf{R}}^{b}_{l^{i}}
(\bm{\zeta}+\bar{\mathbf{p}}) 
+ (\bm{\rho}+\bar{\mathbf{t}}^{b}_{l^{i}})
\\
&\approx
\Big(
\mathbf{I}+\bm{\phi}^{\wedge}
\Big)   
\bar{\mathbf{R}}^{b}_{l^{i}}
(\bm{\zeta}+\bar{\mathbf{p}})
+ 
(\bm{\rho}+\bar{\mathbf{t}}^{b}_{l^{i}}),
\end{aligned}
\end{equation}
where we have kept the first-order approximation of the exponential map. 
If we multiply out the equation and retain only those terms that are first-order in $\bm{\phi}$ or $\bm{\zeta}$, we have
\begin{equation}
\begin{aligned}
\mathbf{y}
\approx
\mathbf{h} + 
\mathbf{H}
\bm{\theta},
\end{aligned}       
\end{equation}
where 
\begin{equation}
\label{equ.theta_definition}
\begin{aligned}
\bm{\theta}
&=
[
\bm{\rho},
\bm{\phi}, 
\bm{\zeta} 
]\\
\mathbf{h}
&=
\bar{\mathbf{R}}^{b}_{l^{i}}\bar{\mathbf{p}} + \bar{\mathbf{t}}\\
\mathbf{H}
&=
[
\mathbf{I}\ \ 
-(\bar{\mathbf{R}}^{b}_{l^{i}}\bar{\mathbf{p}})^{\wedge}\ \ 
\bar{\mathbf{R}}^{b}_{l^{i}}
],
\end{aligned}
\end{equation}
We embody the uncertainties of extrinsics and measurements into $\bm{\theta}$ that is subjected to a zero-mean Gaussian with $9\times9$ covariance
$\bm{\Theta}=\text{diag}(\mathbf{P}, \bm{\Phi}, \mathbf{Z})$.
Linearly transformed by $\bm{\theta}$, $\mathbf{y}$ is a Gaussian variable with mean and covariance as
\begin{equation}
\label{equ.uncertainty_of_each_point}
\begin{aligned}
\bm{\mu}
&\triangleq
E[\mathbf{y}]
=
\mathbf{h}\\
\bm{\Xi}
&\triangleq
E\big[
(\mathbf{y}-\bm{\mu})(\mathbf{y}-\bm{\mu})^{\top}
\big]
=
\mathbf{H}\bm{\Theta}\mathbf{H}^{\top}.
\end{aligned}   
\end{equation}
where we follow \cite{kim2017uncertainty} to use the trace, i.e., $\text{tr}(\bm{\Xi})$, as the criterion to quantify the magnitude of a covariance. 

\begin{figure}
	\centering
	\includegraphics[width=0.48\textwidth]{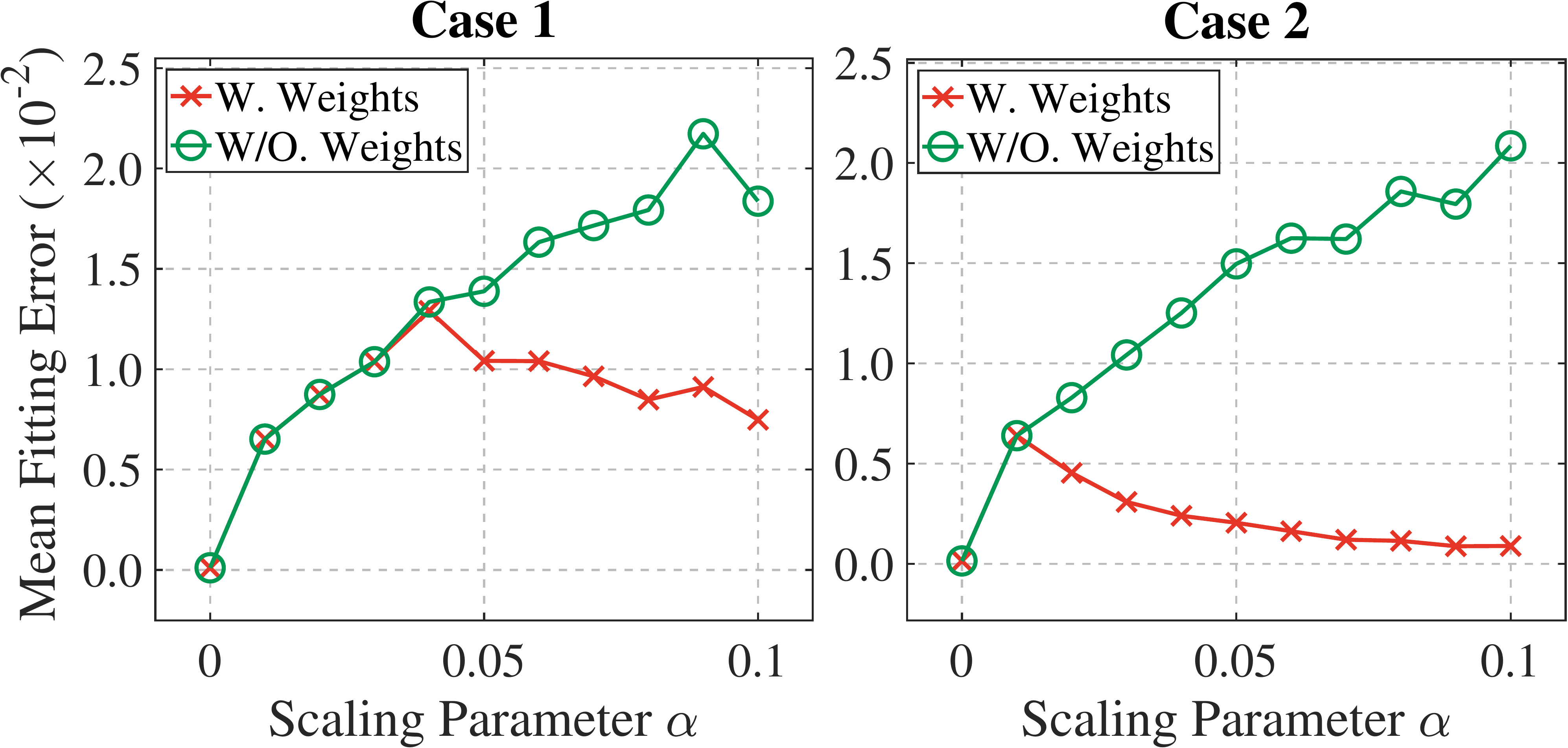}
	\caption{Mean fitting error $e_{\mathbf{x}} = ||\mathbf{x}_{gt} - \mathbf{x}_{est}||$ of methods with weights (use uncertainty prior) or without weights on two planar surface cases. The weighted method does better, and the error does not increase along with $\alpha$.}       
	\label{fig.problem_plane_fitting_experiment}     
	\vspace{-0.5cm}    
\end{figure}

\begin{figure*}
	\centering
	\includegraphics[width=0.9\textwidth]{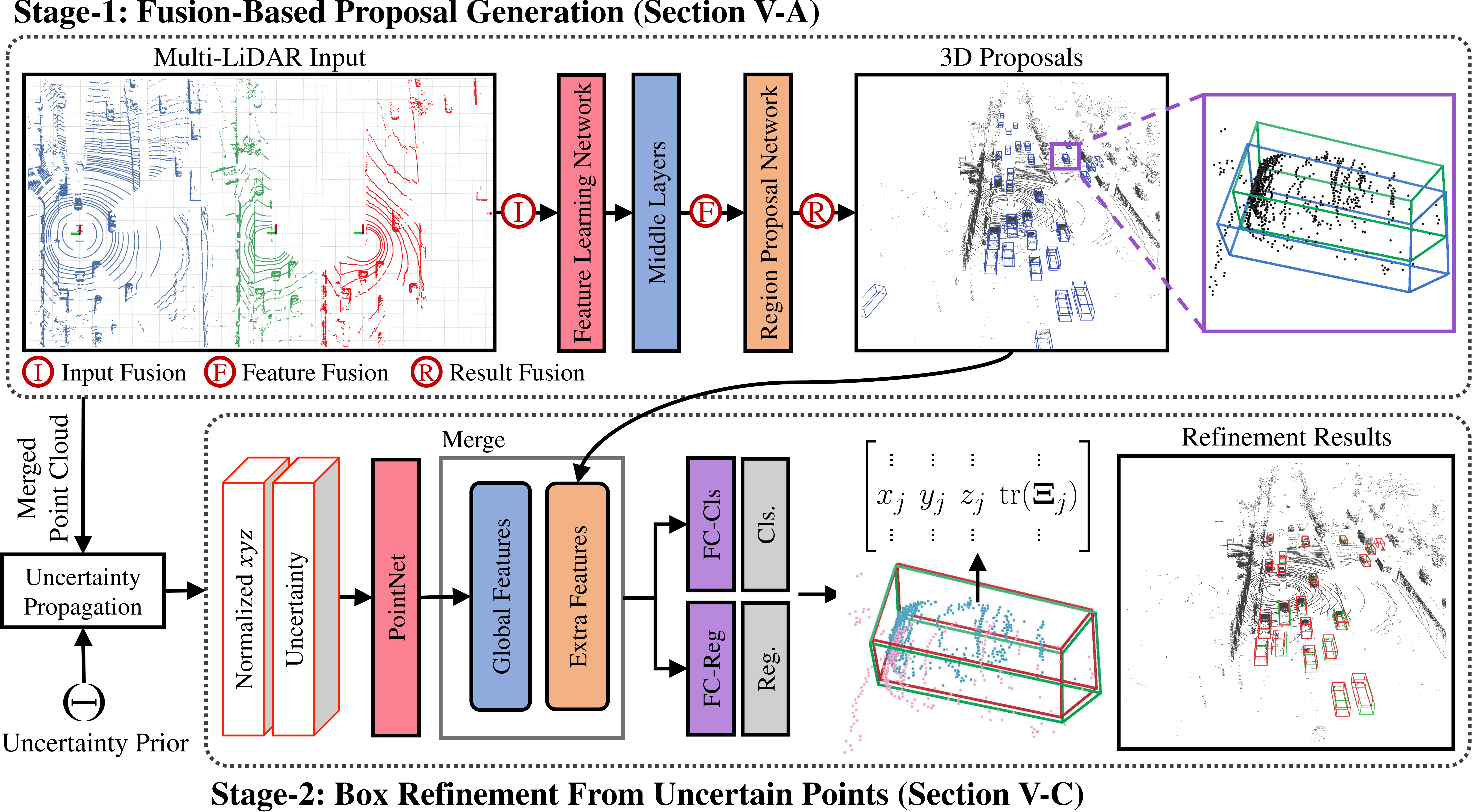}
	\caption{Overview of the proposed detector (\textit{MLOD}) for multi-LiDAR 3D object detection. 
	Red circles: the position where the fusion may be performed;
	Blue cubes: proposals in stage one;
	Red cubes: refinements in stage two;
	Green cubes: ground truths.
	We used the score and parameter of a proposal from the stage-1 network as the extra features. 
	Global features are extracted by PointNet\cite{qi2017pointnet}. They are merged as the input for final classification and regression.
	For simplicity, we consider three LiDARs as an example; however, \textit{MLOD} is extensible to more. 
	}  
	\label{fig.pipeline}    
	\vspace{-0.4cm}    
\end{figure*} 

\subsection{Uncertainty Propagation Example}
\label{sec.problem_propagation}
In this section, a simple example of uncertainty propagation on a point is presented, where we quantify the data noise caused by both extrinsic and measurement perturbation.
The associated covariance is propagated by passing the Gaussian uncertainty from extrinsics and measurements through a transformation.
We use the ground-truth extrinsics having rotation with $[10, 10, 10]^{\circ}$ in roll, pitch, yaw and translation with $[1, 1, 1]m$ along the $x\textendash$, $y\textendash$, and $z\textendash$ axis respectively. Let $\mathbf{p}=[10, 10, 10]^{\top}m$ be a landmark.
$\bm{\Theta}$ is treated as our prior knowledge and considered as a constant matrix\footnote{The measurement noise is found in LiDAR datasheets. 
The extrinsic perturbation typically has the variances around $5cm$ in translation and $5^{\circ}$ in rotation. This is based on our studies on multi-LiDAR systems \cite{jiao2019automatic,liu2020hercules}.}
\begin{equation}
	\bm{\Theta} = 
	\text{diag}^{2}
	\bigg\{
	\frac{1}{20},
	\frac{1}{20},
	\frac{1}{20},
	\frac{1}{10},
	\frac{1}{10},
	\frac{1}{10},	
	\frac{1}{50},
	\frac{1}{50},
	\frac{1}{50}\bigg\},
	\label{equ.problem_theta_define}
\end{equation} 
where the first six diagonal entries can be multiplied by a scaling parameter $\alpha$, allowing us to parametrically increase the extrinsic covariance in experiments. According to \eqref{equ.uncertainty_of_each_point}, we have an approximate expression
\begin{equation}
\begin{aligned}
	\bm{\Xi}
	&\approx
	\alpha\times
    \begin{bmatrix*}[r]
    2.36 & -1.24 & -1.20\\
    -1.24 & 2.41 & -1.18\\
    -1.20 & -1.18 & 2.49\\ 
    \end{bmatrix*},
\end{aligned}	
\end{equation}
where the new position has a high variance around $0.22m$ on each axis even for a small input covariance (i.e., $\alpha=0.02$).
This accuracy is totally unacceptable for autonomous driving.

\subsection{Plane Fitting Experiment}
\label{sec.problem_fitting}
We denote $\mathcal{P}_{m}$ a merged point cloud which is obtained by transforming $\mathcal{P}^{l^{1}}$ and $\mathcal{P}^{l^{2}}$ into $\{\}^{b}$ with ground-truth extrinsics.
We assume that there is a dominant plane in $\mathcal{P}_{m}$, and our task is to estimate the plane coefficients.
They can be obtained by solving a linear system $\mathbf{A}\mathbf{x}=\mathbf{b}$ from $\mathcal{P}_{m}$, where row elements of $\mathbf{A}$ are the point coordinates, and $\mathbf{b}$ is set as an identity vector with least-squares methods. 

Here, we show how the input uncertainty prior $\bm{\Theta}$ can be utilized to acquire better fitting results.
We define $\mathbf{W}$ as a diagonal matrix to weight the linear system. 
For a point $\mathbf{p}_{j}\in\mathcal{P}_{m}$, we propagate its associated covariance $\bm{\Xi}$ using \eqref{equ.uncertainty_of_each_point}. 
The corresponding entry (i.e., $\mathbf{W}_{jj}$) is set as the inverse of $\text{tr}(\bm{\Xi})$.
After that, the fitting problem is turned into a weighted least-squares regression and the optimal results can be obtained as $\mathbf{x}^{*} = (\mathbf{A}^{\top}\mathbf{W}\mathbf{A})^{-1}\mathbf{A}^{\top}\mathbf{W}\mathbf{b}$.

An experiment is conducted on a toy example to compare the performance of two fitting methods (i.e., with or without weights). 
We use the ground-truth extrinsics and the prior covariance in Section \ref{sec.problem_propagation} to sample perturbation with 100 trials.
We also generate $10000$ points according to a planar surface which are subjected to zero-mean Gaussian noise with a standard deviation of $0.02m$ to produce $\mathcal{P}_{m}$. $\mathcal{P}_{m}$ is randomly split into two equal parts to form $\mathcal{P}^{l^{1}}$ and $\mathcal{P}^{l^{2}}$.
At each trial, we evaluate the plane fitting results of each method by comparing them to the ground truth as $e_{\mathbf{x}} = ||\mathbf{x}_{gt} - \mathbf{x}_{est}||$.

The mean fitting error of each method on two different cases over $\alpha\in[0, 0.1]$ is shown in Fig. \ref{fig.problem_plane_fitting_experiment}. 
We see that the weighted least-squares method does better in estimating the plane coefficients, and the mean fitting error does not increase along with $\alpha$.
It shows that the point-wise uncertainty information can be utilized to improve the robustness of algorithms in geometric tasks.
But in practice, the value of $\alpha$ should be carefully set by manual or adaptively obtained from an online method.
Otherwise, a few correct measurements are discarded during operation.

%% file: methodology.tex
\section{Methodology}
\label{sec:methodology}

In this section, we extend our findings from the basic geometric tasks to multi-LiDAR-based 3D object detection. 
The overall structure of the proposed two-stage \textit{MLOD} is illustrated in Fig. \ref{fig.pipeline}.
We adopt SECOND \cite{yan2018second}, which is a sparse convolution improvement of VoxelNet \cite{zhou2018voxelnet}, to generate 3D proposals in the first stage.
\footnote{We use SECOND V1.5 in our experiments: \url{https://github.com/traveller59/second.pytorch}.
Different from the original SECOND \cite{yan2018second}, the FLN of SECOND V1.5 is simplified by computing the mean value of points within each voxel, which consumes less memory.}
It consists of three components:
a feature learning network (\textbf{FLN}) for feature extraction;
middle layers (\textbf{ML}) for feature embedding with sparse convolution;
and a region proposal network (\textbf{RPN}) for box prediction and regression.
Readers are referred to \cite{zhou2018voxelnet, yan2018second} for more details about the network structures.
We first present three species of 3D proposal generation with different multi-LiDAR fusion schemes:
\textit{Input Fusion}, \textit{Feature Fusion} and \textit{Result Fusion}.
Then, we introduce the architecture of our stage-2 network to tackle the extrinsic uncertainty and refine the proposals.

\subsection{Proposal Generation With Three Fusion Schemes}
\label{sec.three_fusion_schemes}
According to stages at which the information from multiple LiDARs is fused, we propose three general fusion schemes, which we call \textit{Input Fusion}, \textit{Feature Fusion}, and \textit{Result Fusion}. These approaches are developed from SECOND with several modifications.

\subsubsection{Input Fusion}
\label{sec.input_fusion}
The fusion of point clouds is performed at the input stage.
We transform raw point clouds perceived by all LiDARs into the base frame to obtain the fused point cloud, and then feed it to the network as the input.
\subsubsection{Feature Fusion}
\label{sec.feature_fusion}
To enhance the feature interaction, the LiDAR data is also fused in the feature level. 
The extracted features from the FLN and ML are transformed into the base frame, and then fused by adopting the maximum value as
\begin{equation}
\begin{split}
\mathcal{F}_{\text{fused}}
=
\mathcal{F}^{l^{1}}\oplus
(\mathbf{T}_{l^{2}}^{b}\mathcal{F}^{l^{2}})\oplus    
\dots \oplus
(\mathbf{T}_{l^{I}}^{b}\mathcal{F}^{l^{I}}),         
\end{split}
\end{equation}
where $\mathcal{F}^{l^{i}}$ are the extracted features, $\oplus$ denotes the max operator, and $I$ is the number of LiDARs.

\subsubsection{Result Fusion}
\label{sec.result_fusion}
The result fusion takes the box proposals and the associated points as inputs, and produces a set of boxes with high scores.
We transform all box proposals into the base frame, and then filter them as
\begin{equation}
\begin{split}
\mathcal{B}_{\text{fused}}
=
\mathcal{B}^{l^{1}}\oplus
(\mathbf{T}_{l^{2}}^{b}\mathcal{B}^{l^{2}})\oplus    
\dots \oplus
(\mathbf{T}_{l^{I}}^{b}\mathcal{B}^{l^{I}}),
\end{split}
\end{equation}
where $\oplus$ denotes the non-maximum suppression (NMS) on 3D intersection-over-Union (IoU).
After transformation, each object is associated with several candidate boxes, and these boxes with low confidences are filtered by the NMS.

\subsection{Box Refinement From Uncertain Points}
\label{sec.box_refinemnet_from_uct_pts}
Although the stage-1 fusion-based network generates promising proposals, its capability in handling uncertain data uncertainty is fragile.
Therefore, we propose a stage-2 module to improve the robustness of \textit{MLOD}. 
A straight-forward idea is to eliminate the highly uncertain points according to their associated covariances. 
But this method is sensitive to a pre-set threshold. 
Inspired by \cite{shi2019pointrcnn},
it is more promising that training a neural network to embed features and refine the proposals with an awareness of uncertainty.

We thus use a deep neural network to deal with this problem. 
The network takes a series of points of each proposal (a $2m$ margin along $x\textendash$, $y\textendash$, and $z\textendash$ axes is expanded) generated by the stage-1 network, and refines the stage-1 proposals.
In addition to three-dimensional coordinates, each point is also embedded with its uncertain quantity. 
As defined in \eqref{equ.uncertainty_of_each_point}, we use the trace, i.e., $\text{tr}(\bm{\Xi})$, to quantify the uncertainties.
We employ PointNet \cite{qi2017pointnet} as the backbone to extract global features.
We also encode extra features, including the scores and parameters of each proposal, to concatenate with the global features.
At the end of the network, fully connected layers are used to provide classification scores and refinement results with two heads.
To reduce the variance of input, we normalize the coordinates of each point of a proposal as 
\begin{equation}
\begin{aligned}
	\mathbf{p}_{\text{n}}
	&= 
	\mathbf{T}\mathbf{p}
	 \odot \mathbf{s},
\end{aligned}
\end{equation} 
where
\begin{equation}
\begin{aligned}
	\mathbf{T} 
	& = 
	\begin{bmatrix}
		cos(\gamma) &  sin(\gamma)& 0 & -x\\
		-sin(\gamma) & cos(\gamma) &  0& -y\\
		0 & 0 & 1 & -z\\
		0 & 0 & 0 & 1
	\end{bmatrix} \\
	\mathbf{s} 
	& = 
	\begin{bmatrix}
		1/l & 1/w & 1/h & 1
	\end{bmatrix}^{\top},
\end{aligned}
\end{equation}
where $\mathbf T$ and $\mathbf s$ are defined in terms of the parameters of a proposal $\mathbf b = [c, x, y, z, l, w, h, \gamma]$, $\odot$ is the Hadamard product, and both $\mathbf{p}_{\text{n}}$ and $\mathbf p$ are represented in homogeneous coordinates.
The output of the classification head is a binary variable indicating the probability of objectiveness.
We regress residuals of the bounding box based on the proposal 
instead of directly regressing the final 3D bounding box.
\begin{figure}
	\centering
	\includegraphics[width=0.45\textwidth]{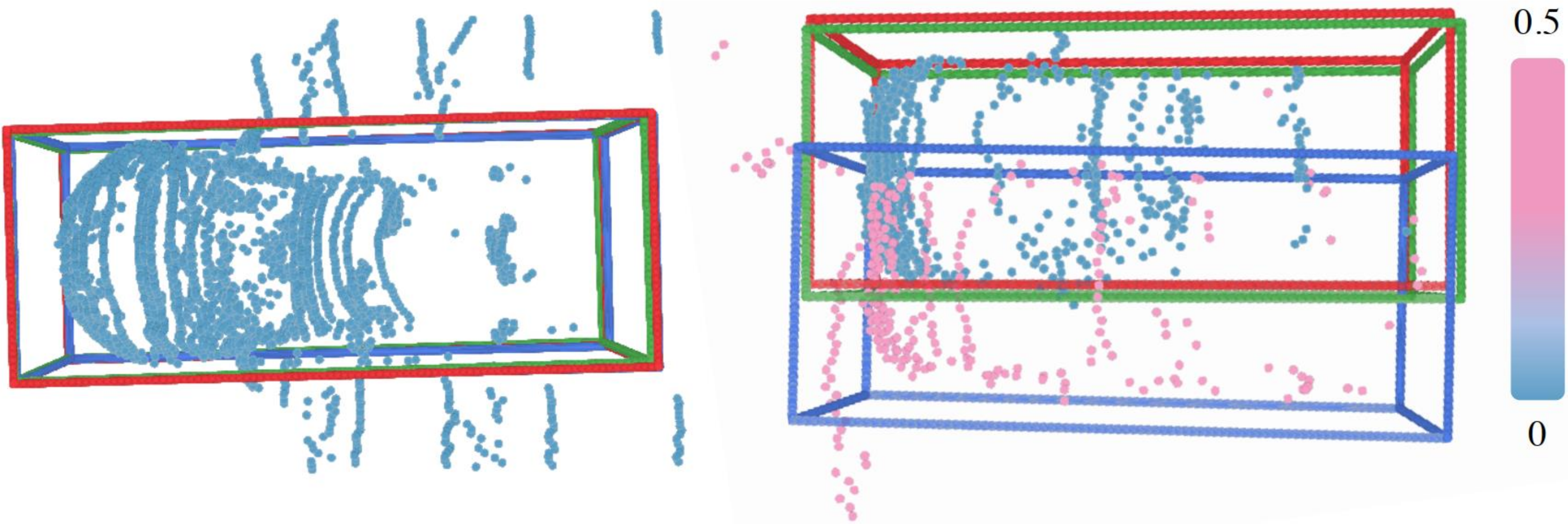}
	\caption{Illustration of the \textit{MLOD} results on the proposals and merged point cloud given by the stage-1 network.
		The color of each point represents its uncertainty quantity defined as the trace of the associated covariance. Blue to pink color indicates low to high uncertainty.
		The estimated and ground-truth bounding boxes are also marked with different colors: blue as the proposals in stage one, 
		red as refinements in stage two, and green as ground truths.
	}
	\label{fig.mlod_refinement}
	\vspace{-0.5cm}
\end{figure}
The regression target of the stage-2 network is 
\begin{equation}
\begin{aligned}
	\mathbf{u}_{i} \triangleq 
	&\ \big[\frac{x - x_p}{l_p}, 
	 \frac{y - y_p}{w_p},
	 \frac{z - z_p}{h_p}, 
	 \frac{l - l_p}{l_p}, \\
	&
	 \frac{w - w_p}{w_p},
	 \frac{h - h_p}{h_p},
	 \sin(\gamma - \gamma_p),
	 \cos(\gamma - \gamma_p)\big],
\end{aligned}
\label{equ.regression_output}
\end{equation}
where $\mathbf{u}_{i} \in\mathbb{R}^{8}$ are regression outputs for the $i^{th}$ positive proposal respectively. 
We adopt the classification and regression loss which are defined in  \cite{yun2019focal} as 
\begin{equation}
\mathcal{L}_{\text{cls}} 
=
\sum_{i}^{}f_{\text{cls}}(p_{i}),\ \ \ 
\mathcal{L}_{\text{reg}}
=
\sum_{i}^{}f_{\text{reg}}(\mathbf{u}_{i}, \mathbf{u}_{i}^{*}),
\label{equ.loss_function}
\end{equation}
where $p_{i}$ represents the posterior probability of objectiveness, 
$\mathcal{L}_{\text{cls}}$ is defined as the classification loss, $f_{\text{cls}}(\cdot)$ denotes the focal loss \cite{lin2017focal}, $\mathcal{L}_{\text{reg}}$ is defined as the normalized regression loss, and $f_{\text{reg}}(\cdot)$ denotes the smooth-L1 loss.
Since we have the observation that the regression outputs $\mathbf{u}_{i}$ should not be large when the uncertainty is small,
we add a regularization term to penalize large $\mathbf{u}_{i}$ in the low uncertainty cases:
\begin{equation}
\begin{split}
\mathcal{L} _{\text{uct}}
= 
\sum_{i}^{}
\exp
\bigg(1 - f\big[\underset{j}{\textit{MAX}}\ \text{tr}(\bm{\Xi}_{j})\big]\bigg)
\cdot
||\mathbf{u}_{i}^{'}||_{2},
\end{split}
\label{equ.uct_loss_function}
\end{equation}
where 
$\text{tr}(\bm{\Xi}_{j})$ is the point-wise uncertainty of the $j^{th}$ point within the $i^{th}$ proposal,
$\textit{MAX}$ is the max operator,
$\mathbf{u}_{i}^{'}$ is the regressed residual without the last cosine term for the the $i^{th}$ proposal,
and $f(\cdot)$ clamps a value into the range $[10^{-3},0.5]$ to stabilize training.
Finally, we define the loss function for training the stage-2 network as
\begin{equation}
	\mathcal{L} 
	=
	\mathcal{L}_{\text{reg}}
	+ \eta \mathcal{L}_{\text{cls}} 
	+ \lambda \mathcal{L}_{\text{uct}}\ ,
\end{equation}
where $\eta$ and $\lambda$ are hyper-parameters to balance the weight of the classification loss and the uncertainty regularizer. 
We use $\eta=2$ and $\lambda=0.005$ in our experiments.

Fig. \ref{fig.mlod_refinement} visualizes the refinement results of our stage-2 network, which shows that our method produces similar results to the ground truths even if many points are uncertain.
More quantitative examples are provided in Section \ref{sec:experiment_ruc}.

%% file: experiment.tex
\section{Experiment}
\label{sec:experiment}
In this section, we evaluate our proposed \textit{MLOD} on LYFT multi-LiDAR dataset \cite{lyft2019} in terms of accuracy and robustness under different levels of extrinsic perturbation. 
In particular, we aim to answer the following questions: 
\begin{enumerate}
	\item Can a multi-LiDAR object detector perform better accuracy than single-LiDAR methods? 
	\item Can \textit{MLOD} improve the robustness of a multi-LiDAR object detector under extrinsic perturbation? 
\end{enumerate}

\subsection{Implementation Details}
\subsubsection{Dataset} 
The LYFT dataset \cite{lyft2019} provides a large amount of data collected in a variety of environments for the task of 3D object detection.
Three $40$-beam and calibrated Hesai LiDARs are mounted on the top, front-left and front-right position of the vehicle platform.
The top LiDAR is set as the primary LiDAR, which is denoted by $l^{1}$. The left and right LiDARs are auxiliary LiDARs, which are denoted by $l^{2}$ and $l^{3}$ respectively.
This setting is according to LiDAR's field of views (FOV).
We select all multi-LiDAR data samples for our experiment.
The data contain $4031$ samples, $2500$ of which are for training as well as validation and $1531$ of which are for testing.
The testing scenes are different from those in the training and validation sets.

\subsubsection{Metric}
Following the KITTI evaluation metrics \cite{geiger2013vision},
we compute the average precision on 3D bounding boxes ($\text{AP}_{\text{3D}}$) to measure the detection accuracy.
We compute the $\text{AP}_{\text{3D}}$ for $360^{\circ}$ around the vehicle instead of evaluating only the $90^{\circ}$ front view.
According to the distance of objects, 
we set three different evaluation difficulties: \textit{easy} ($<20m$), \textit{moderate} ($<30m$), and \textit{hard} ($<50m$).

\subsubsection{Extrinsic Perturbation Injection}
\label{sec.extrinsic_per_inject}
We inject extrinsic perturbation on the original LYFT dataset to generate another set of data for robustness evaluation.
We take $\bm{\Theta}$ defined in \eqref{equ.problem_theta_define} and adjust $\alpha\in[0,0.1]$ with a $0.02$ interval to simulate different levels of perturbation. 
To remove the effects from outliers, we bound the sampled perturbation within the $\sigma$ position.
Although the maximum value of $\alpha$ is small, but the effect of input perturbation on points is obvious.
The noisy extrinsics are obtained by adding the sampled perturbation to the ground-truth extrinsics according to \eqref{equ.translation_with_uncertainty} and \eqref{equ.rotation_with_uncertainty}.

\subsubsection{Case Declaration}
We declare all cases of our methods which are tested in the following experiments as
\begin{itemize}
    \item \textit{LiDAR-Top}, \textit{LiDAR-Left}, \textit{LiDAR-Right}: 
    single-LiDAR objeect detectors based on SECOND with the input data which are captured by the top LiDAR, front-left LiDAR and front-right LiDAR, respectively.
    \item \textit{Input Fusion}, \textit{Feature Fusion}, \textit{Result Fusion}:
    multi-LiDAR objeect detetors with different fusion schemes, which are introduced in Section \ref{sec.three_fusion_schemes}.
    \item \textit{MLOD-I}, \textit{MLOD-F}, \textit{MLOD-R}:
	\textit{MLOD} with \textit{Input Fusion}, \textit{Feature Fusion}, \textit{Result Fusion} as its stage-1 network respectively, which are described in Section \ref{sec.box_refinemnet_from_uct_pts}.
	\item \textit{MLOD-I (OC)}, \textit{MLOD-F (OC)}, \textit{MLOD-R (OC)}: Variants of \textit{MLOD} with an online calibration method to reduce extrinsic perturbation. 
	This method is implemented with a point-to-plane ICP approach \cite{pomerleau2013comparing}. 
\end{itemize}

\subsubsection{Training of Stage-One Network}
We follow \cite{yan2018second} to train the stage-1 network.
During training, we conduct dataset sampling as in \cite{yan2018second}
and an augmentation of random flips on the $y\textendash$ axis as well as translation sampling within $[-0.5, 0.5]m$, $[-0.5, 0.5]m$, $[-0.3, 0.3]m$ on the $x\textendash$, $y\textendash$, and $z\textendash$ axes respectively,
as well as rotation around the $z\textendash$ axis between $[-5, 5]^{\circ}$.
In each epoch, the augmented data accounts for $30\%$ of the whole training data. 
In \textit{Result Fusion} and \textit{Feature Fusion},
we directly take the calibrated and merged point clouds from different LiDARs as input.
In \textit{Result Fusion}, we calibrate all the proposal generators with temperature scaling \cite{guo2017calibration}.

\subsubsection{Training of Stage-Two Network}
To train our stage-2 network, we use all cases of the  well-trained stage-1 networks to generate proposals.
We use the training set to create proposals and inject uncertainties 
by selecting different $\alpha$ like the above section.
In addition, we force the ratio between uncertainty-free samples and uncertain samples to be about $1$:$1.5$ to stabilize the training process.
A proposal is considered to be positive if its maximum IoU with ground-truth boxes is above $0.6$,
and is treated as negative if its maximum 3D IoU is below $0.45$.
During training, we conduct data augmentation of random flipping, 
scaling with a scale factor sampled from $[0.95, 1.05]$, 
translation along each axis between $[-0.1, 0.1]m$
and rotation around the $z$- axis between $[-3, 3]^{\circ}$.
We also randomly sample $1024$ points within the proposals to increase their diversity.

\begin{table}[]
	\centering
	\caption{Average Precision on the multi-LiDAR LYFT test set}
	\renewcommand\arraystretch{1.1}
	\begin{tabular}{ccccccc}
		\toprule[0.03cm]
		\multirow{2}{*}{Case} & 
		\multicolumn{3}{c}{$\text{AP}_{\text{3D}}\ \text{IoU}\geqslant0.7$}     & \multicolumn{3}{c}{$\text{AP}_{\text{3D}}\ \text{IoU}\geqslant0.5$}    \\
		& easy          & mod.          & hard          & easy          & mod.          & hard          \\ 
		\hline
		\toprule[0.03cm]
		\textit{LiDAR-Top}      & \textbf{62.7} & \textbf{52.9} & \textbf{37.8} & \textbf{89.8} & \textbf{80.1} & \textbf{61.8} \\
		\textit{LiDAR-Left}     & 41.1          & 37.0          & 27.5          & 62.6          & 53.9          & 43.1          \\
		\textit{LiDAR-Right}    & 40.9          & 31.5          & 23.4          & 53.9          & 52.4          & 35.5          \\ 
		\toprule[0.01cm]
		\textit{Input Fusion}   & 71.9          & 61.8          & 45.0          & 89.4          & 87.9          & \textbf{61.9} \\
		\textit{MLOD-I}         & \textbf{72.4} & \textbf{62.7} & \textbf{45.6} & \textbf{89.6} & \textbf{88.3} & 61.8          \\ 
		\toprule[0.01cm]
		\textit{Feature Fusion} & 71.1          & 60.2          & 44.1          & 89.6          & 87.4          & \textbf{61.6} \\
		\textit{MLOD-F}         & \textbf{72.0} & \textbf{61.8} & \textbf{44.9} & \textbf{89.7} & \textbf{88.2} & \textbf{61.6} \\ 
		\toprule[0.01cm]		
		\textit{Result Fusion}  & 71.4          & 61.6			& 44.5          & 89.5 	        & 87.7          & \textbf{62.0} \\
		\textit{MLOD-R}         & \textbf{72.1} & \textbf{62.3} & \textbf{45.2} & \textbf{89.6} & \textbf{88.0} & 61.7          \\ 
		\toprule[0.03cm]
	\end{tabular}
	\label{tab.lyft_acc}
	\vspace{-0.4cm}	
\end{table}

\begin{table*}[]
	\centering
	\caption{Mean and Variance of Accuracy ($\text{AP}_{\text{3D}}\ \text{IoU}\geqslant0.7$) Under Different Level of the Extrinsic Perturbation.}
    \renewcommand\tabcolsep{2.5pt}
    \renewcommand\arraystretch{1.1}
\begin{tabular}{@{}ccccccccccccc@{}}
   	\toprule[0.03cm]
   	\multirow{2}{*}{Cases} & \multicolumn{3}{c}{$\alpha = 0$}              & \multicolumn{3}{c}{$\alpha = 0.02$}                                & \multicolumn{3}{c}{$\alpha = 0.04$}                                & \multicolumn{3}{c}{$\alpha = 0.1$}                                 \\
   	& \textit{easy}          & \textit{mod.}          & \textit{hard}          & \textit{easy}                 & \textit{mod.}                 & \textit{hard}                 & \textit{easy}                 & \textit{mod.}                 & \textit{hard}                 & \textit{easy}                 & \textit{mod.}                 & \textit{hard}                 \\ \toprule[0.03cm]
   	\textit{Top LiDAR}           & 63.3          & 53.8          & 38.2         & 63.3                 & 53.8                 & 38.2                 & 63.3                 & 53.8                 & 38.2                 & 63.3                 & 53.8                 & 38.2                 \\ \toprule[0.03cm]
   	\textit{Input Fusion}        & \textbf{71.2} & 62.3          & 45.4         & 64.7 $\pm$ 2.5 & 60.8 $\pm$ 0.4          & 44.4 $\pm$ 0.2          & \textbf{63.0} $\pm$ 0.5 & 53.9 $\pm$ 2.1          & \textbf{38.0} $\pm$ 0.2 & 60.9 $\pm$ 1.0          & 51.0 $\pm$ 0.6          & \textbf{36.1} $\pm$ 0.4 \\
   	\textit{MLOD-I}              & \textbf{71.2} & \textbf{62.8} & \textbf{46.1} & \textbf{66.3} $\pm$ 3.6 & \textbf{61.4} $\pm$ 0.4 & \textbf{44.8} $\pm$ 0.2 & 62.8 $\pm$ 0.4          & \textbf{56.3} $\pm$ 3.5 & 37.7 $\pm$ 0.2          & \textbf{61.6} $\pm$ 0.9 & \textbf{51.5} $\pm$ 0.6 & 35.8 $\pm$ 0.5          \\ \toprule[0.01cm]
   	\textit{Input Fusion (OC)}   & -             & -             & -            & 66.8 $\pm$ 4.0          & 61.2 $\pm$ 0.5          & 44.5 $\pm$ 0.2          & 63.4 $\pm$ 0.4          & 60.0 $\pm$ 0.5          & 42.8 $\pm$ 2.2          & 62.4 $\pm$ 0.6          & 52.3 $\pm$ 0.4          & \textbf{37.2} $\pm$ 0.2 \\
   	\textit{MLOD-I (OC)}         & -             & -             & -            & \textbf{67.6} $\pm$ 4.0 & \textbf{61.5} $\pm$ 0.5 & \textbf{44.9} $\pm$ 0.4 & \textbf{64.3} $\pm$ 2.5 & \textbf{60.4} $\pm$ 0.6 & \textbf{43.1} $\pm$ 2.5 & \textbf{62.9} $\pm$ 0.4 & \textbf{52.9} $\pm$ 0.3 & 36.8 $\pm$ 0.3          \\ \toprule[0.03cm]
   	
   	\textit{Feature Fusion}      & 71.2          & 59.2          & 42.9         & 63.9 $\pm$ 2.5          & 59.2 $\pm$ 0.9          & 42.9 $\pm$ 0.9          & 62.9 $\pm$ 0.4          & 58.6 $\pm$ 2.5          & 41.1 $\pm$ 3.3          & 61.9 $\pm$ 0.6          & 51.1 $\pm$ 1.0          & 35.7 $\pm$ 0.7          \\
   	\textit{MLOD-F}              & \textbf{71.5} & \textbf{61.8} & \textbf{45.2} & \textbf{65.4} $\pm$ 3.3 & \textbf{60.9} $\pm$ 0.4 & \textbf{44.5} $\pm$ 0.2 & \textbf{63.7} $\pm$ 0.5 & \textbf{59.6} $\pm$ 2.0 & \textbf{42.1} $\pm$ 2.8 & \textbf{63.7} $\pm$ 0.5 & \textbf{53.9} $\pm$ 0.5 & \textbf{37.7} $\pm$ 0.4 \\  \toprule[0.01cm]
   	\textit{Feature Fusion (OC)} & -             & -             & -            & 66.2 $\pm$ 3.7          & 59.5 $\pm$ 0.8          & 43.5 $\pm$ 0.9          & 63.8 $\pm$ 2.2          & 59.6 $\pm$ 0.7          & 43.6 $\pm$ 0.7          & 63.3 $\pm$ 2.4          & 54.3 $\pm$ 3.0          & 37.9 $\pm$ 2.0          \\ 
   	\textit{MLOD-F (OC)}         & -             & -             & -            & \textbf{67.7} $\pm$ 3.8 & \textbf{61.2} $\pm$ 0.4 & \textbf{44.7} $\pm$ 0.2 & \textbf{65.3} $\pm$ 3.2 & \textbf{60.4} $\pm$ 0.4 & \textbf{44.1} $\pm$ 0.2 & \textbf{64.9} $\pm$ 2.1 & \textbf{58.9} $\pm$ 2.9 & \textbf{38.2} $\pm$ 0.2 \\ \toprule[0.03cm]
   	
   	\textit{Result Fusion}       & 70.4          & 62.1          & 44.0         & 64.5 $\pm$ 2.3          & 60.1 $\pm$ 0.4          & 43.4 $\pm$ 0.2          & 63.0 $\pm$ 0.5          & 55.5 $\pm$ 2.9          & 38.1 $\pm$ 0.2          & 61.1 $\pm$ 1.0          & 51.5 $\pm$ 0.5          & 36.8 $\pm$ 0.3          \\
   	\textit{MLOD-R}              & \textbf{70.7} & \textbf{62.4} & \textbf{45.1}& \textbf{65.7} $\pm$ 2.8 & \textbf{60.9} $\pm$ 0.4 & \textbf{44.2} $\pm$ 0.2 & \textbf{63.1} $\pm$ 0.5 & \textbf{59.3} $\pm$ 0.5 & \textbf{39.4} $\pm$ 2.5 & \textbf{62.6} $\pm$ 0.9 & \textbf{52.9} $\pm$ 0.5 & \textbf{37.0} $\pm$ 0.3 \\
   	 \toprule[0.01cm]
   	\textit{Result Fusion (OC)}  & -             & -             & -            & 66.5 $\pm$ 3.7          & 60.4 $\pm$ 0.6          & 43.5 $\pm$ 0.2          & 64.0 $\pm$ 2.2          & 59.3 $\pm$ 0.4          & 42.0 $\pm$ 1.9          & 62.5 $\pm$ 0.6          & 52.5 $\pm$ 0.4          & \textbf{37.5} $\pm$ 0.2 \\
   	\textit{MLOD-R (OC)}         & -             & -             & -            & \textbf{66.8} $\pm$ 3.8 & \textbf{61.1} $\pm$ 0.5 & \textbf{44.3} $\pm$ 0.4 & \textbf{65.1} $\pm$ 3.2 & \textbf{59.9} $\pm$ 0.6 & \textbf{43.6} $\pm$ 0.4 & \textbf{63.3} $\pm$ 0.5 & \textbf{53.4} $\pm$ 0.3 & \textbf{37.5} $\pm$ 0.2 \\ \toprule[0.03cm]
\end{tabular}
\label{tab.robust_test}
\vspace{-0.3cm}
\end{table*}

\subsection{Results on the Multi-LiDAR LYFT Dataset}
\label{subsec.exp_accuracy}
We evaluate both single-LiDAR and multi-LiDAR detectors on the LYFT test set, as reported in Tab. \ref{tab.lyft_acc}.
In this experiment, we do not consider any extrinsic perturbation between LiDARs.
\textit{LiDAR-Top} performs the best among all single-LiDAR detectors.
This is because the top LiDAR has a complete $360^{\circ}$ horizontal field of view (FOV). The left and right LiDARs, which are partially blocked by the vehicle, only have a $225^{\circ}$ horizontal FOV.
With sufficient measurements and decreasing occlusion areas, all multi-LiDAR detectors outperform \textit{LiDAR-Top}, and the accuracy improvement gains up to $9.8\text{AP}$.
Tab. \ref{tab.lyft_acc} also shows that all \textit{MLOD} variants perform better or comparable to their one-stage counterparts.
This proves that our stage-2 network refines the stage-1 proposals in perturbation-free cases.

\subsection{Robustness Under Extrinsic Perturbation}
\label{sec:experiment_ruc}

\begin{figure*}[h]
	\centering
	\includegraphics[width=0.82\textwidth]{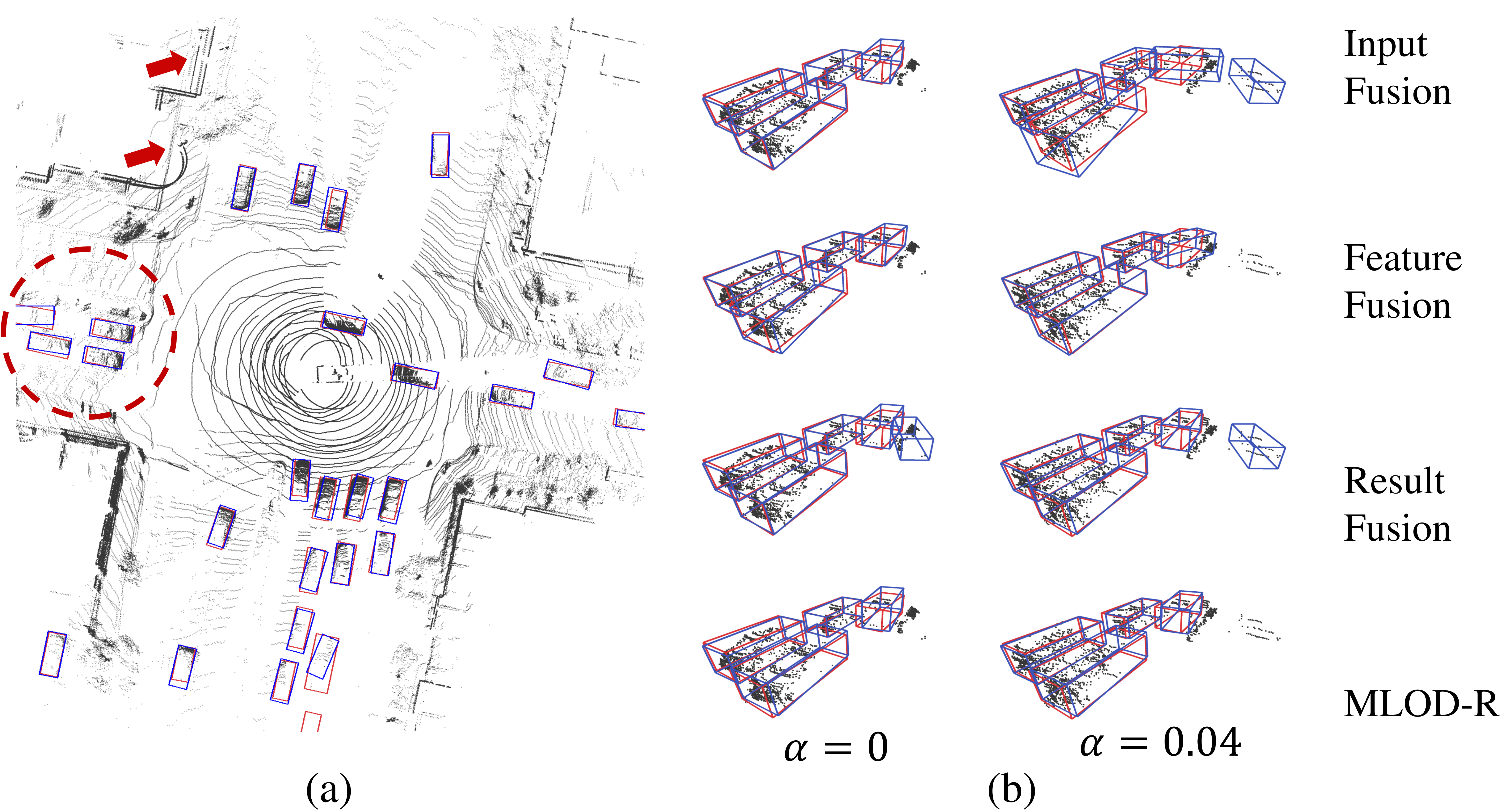}
	\caption{Visualization of the fusion schemes results
    (For comparison, here we sample the extrinsic perturbation for all the ($\alpha=0.04$) cases at the $\sigma$ position according to $\bm{\Theta}$.
    (a) Bird's-eye view of \textit{MLOD}'s detection results when $\alpha=0.04$.
    The extrinsic perturbation is observed from where the red arrows indicate.
    (b) A close-up view of results estimated by different fusion schemes within the red circle. Left to right: $\alpha=0$, $\alpha=0.04$.
    \textit{Input Fusion}, \textit{Feature Fusion} and \textit{Result Fusion} suffer from false positives or inaccurate boxes caused by extrinsic perturbation and inaccurate box location.
    Compared with its counterpart (\textit{Result Fusion}), \textit{MLOD-R} eliminates the false positives and refines the 3D boxes.
	}
	\label{fig.qual_vis}
	\vspace{-0.3cm}
\end{figure*}

In this section, we evaluate the robustness of our proposed methods under extrinsic perturbation.
Three levels of perturbation are performed: no perturbation ($\alpha=0$), moderate perturbation ($\alpha=0.02$, $\alpha=0.04$) and high perturbation ($\alpha=0.1$).
An example is displayed in Fig. \ref{fig.qual_vis}(a) ($\alpha=0.04$), where massive noisy points ($50 cm$ misplacement) appear at $35 m$ away. This is a typical phenomenon of extrinsic perturbation in multi-LiDAR systems.
We randomly select $300$ data samples from the test set with $10$ trials to conduct evaluations at each level.

In Tab. \ref{tab.robust_test}, the detection results in terms of the means and variances as $\text{AP}_{\text{3D}} (\text{IoU}\geqslant0.7)$ are detailed.
We see that all variants of \textit{MLOD} (\textit{MLOD-I}, \textit{MLOD-F}, \textit{MLOD-R}) perform better or comparable to their one-stage counterparts.
Online calibration baselines  (\textit{Input Fusion (OC)}, \textit{Feature Fusion (OC)}, \textit{Result Fusion (OC)}) 
demonstrate better results than those without online calibration.
With the assistance of \textit{MLOD}, their performance is further enhanced.
\textit{Feature Fusion} and its variants perform better than the others.
We explain that \textit{Feature Fusion} fuses data in a high
dimensional embedding space, and each feature is related to
an area of the cognitive region.
Regarding \textit{LiDAR-Top}, \textit{MLOD}'s variants also perform better or comparable results.
A qualitative result is illustrated in Fig. \ref{fig.qual_vis}, and more results are shown in the supplementary materials.

\begin{figure}[]
	\centering
	\subfigure[The details of PointNet \cite{qi2017pointnet}.]    
	{\label{fig.detail_pointnet}\centering\includegraphics[width=0.42\textwidth]{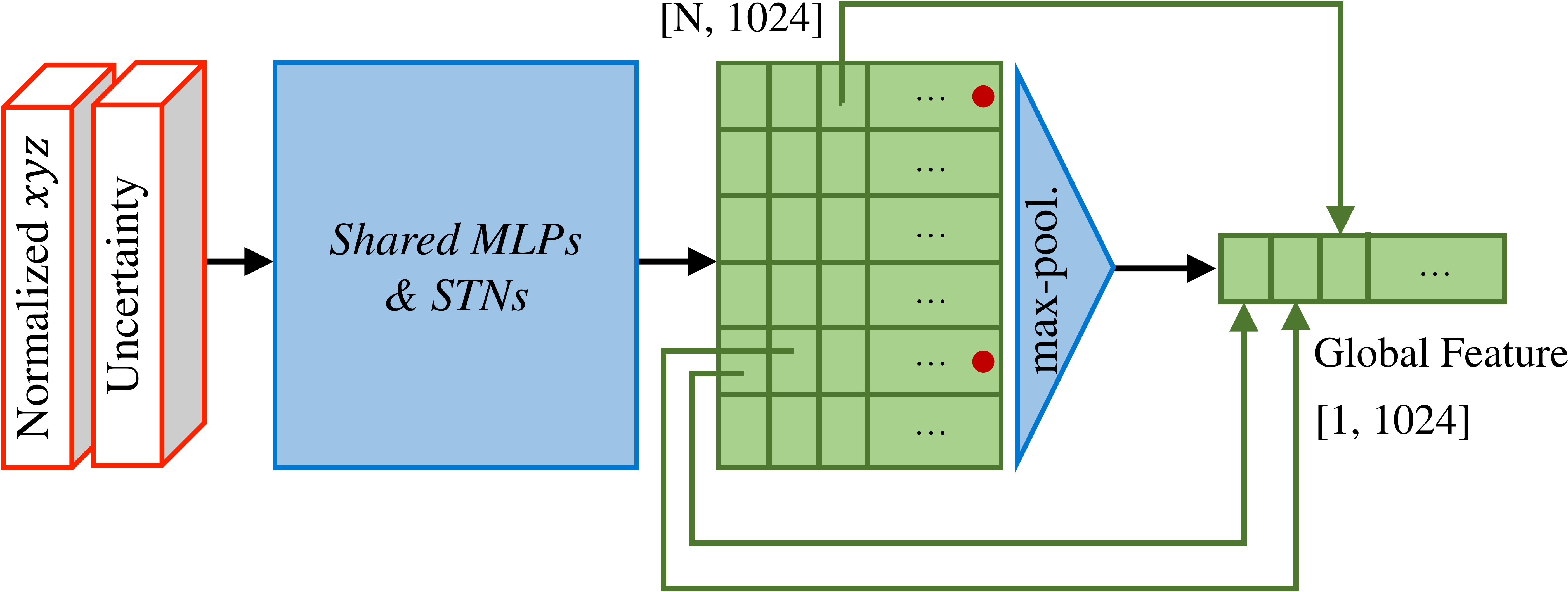}}
	\subfigure[Histogram of proportional differences between input uncertain points and active uncertain points.]    
	{\label{fig.proportion}\centering\includegraphics[width=0.46\textwidth]{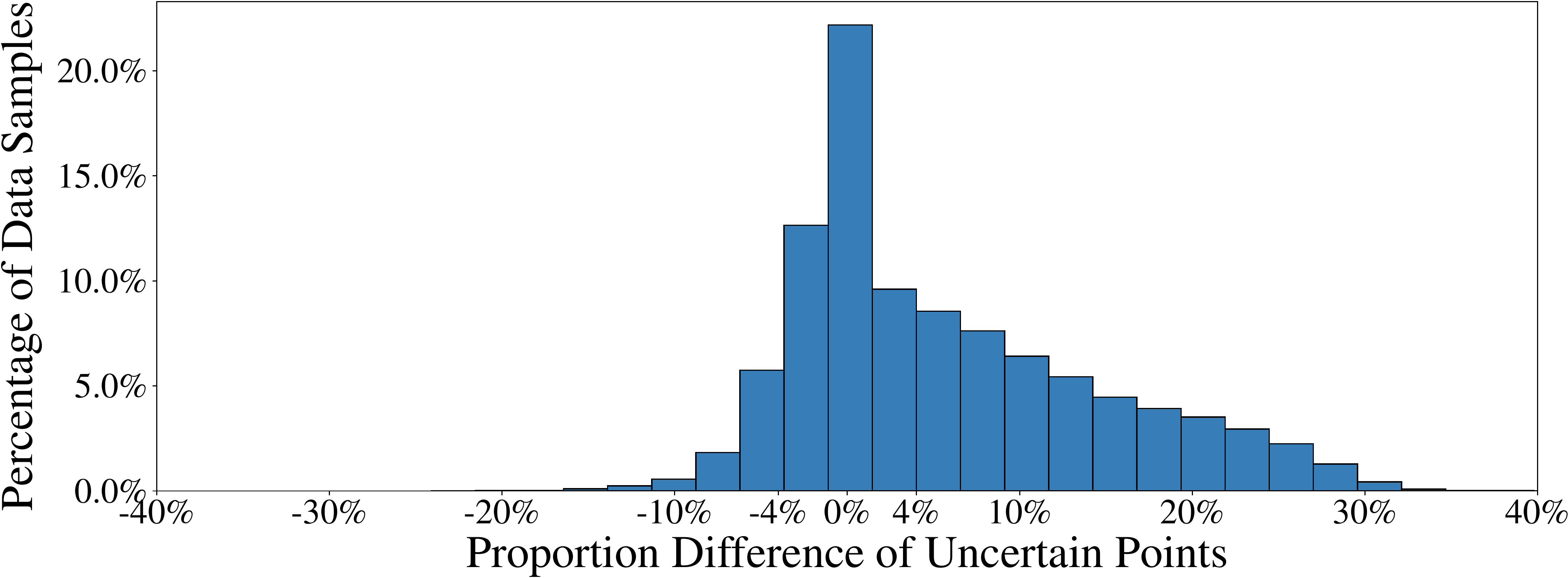}}
	\caption{
	(a) Active points (Red dotted) form the global features after applying the max-pooling symmetric function.
	(b) The proportional differences larger than $0.04$ ($4\%$) occupy about $44.0\%$, and those less than $-0.04$ ($-4\%$) occupy about $8.5\%$. 
	Less uncertain points exist in the active points.}
	\label{fig.mlod_pointnet_detail}
	\vspace{-0.4cm}
\end{figure}

We also conduct a sensitivity analysis of extrinsic uncertainty with a fixed $\alpha$ as input in the supplementary material. The same conclusion still holds.
To further explore the reason behind the performance of \textit{MLOD},
we investigate the characteristics of active points. These points are activated by PointNet before the max-pooling layer (Fig. \ref{fig.detail_pointnet}). 
We compute the difference between the input points and active points on the proportion of uncertain points\footnote{The proportion of the uncertain points is defined as $|\mathcal{P}_{u}|/|\mathcal{P}|$, where $\mathcal{P}_{u}\in\mathcal{P}$ is the set of uncertain points with the uncertain quantity $>0.05$.} for each sample.
Fig. \ref{fig.proportion} plots a histogram of the proportional difference overall input samples.
It shows that the network deactivates highly uncertain points,
and explain why \textit{MLOD} is robust.

%% file: conclusion.tex
\section{Conclusion}
\label{sec:conclusion}

In this paper, 
we analyze the extrinsic perturbation effect on multi-LiDAR-based 3D object detection.
We propose a two-stage network to both fuse the data from multiple LiDARs and handle extrinsic perturbation after data fusion.
We conduct extensive experiments on a real-world dataset and discuss the results in different levels of extrinsic perturbation.
In the perturbation-free situation,
we show the multi-LiDAR fusion approaches obtain better accuracy than single-LiDAR detectors. 
Under extrinsic perturbation, \textit{MLOD} performs great robustness with the assistance of the uncertainty prior.
A future direction concerns the combination with sensors in various modalities, e.g., LiDAR-camera-radar setups, for developing a more accurate object detector.